\documentclass[letterpaper]{article} % DO NOT CHANGE THIS
\usepackage[preprint]{aaai2027}  % arXiv preprint: suppress the AAAI copyright notice
\usepackage[hyphens]{url}  % DO NOT CHANGE THIS
\usepackage{graphicx} % DO NOT CHANGE THIS
\usepackage{natbib}  % DO NOT CHANGE THIS AND DO NOT ADD ANY OPTIONS TO IT
\usepackage{caption} % DO NOT CHANGE THIS AND DO NOT ADD ANY OPTIONS TO IT
\usepackage{algorithm}
\usepackage{algorithmic}
\usepackage{newfloat}
\usepackage{listings}
\DeclareCaptionStyle{ruled}{labelfont=normalfont,labelsep=colon,strut=off} % DO NOT CHANGE THIS
\floatstyle{ruled}
\newfloat{listing}{tb}{lst}{}
\floatname{listing}{Listing}

\usepackage{booktabs}
\usepackage{amssymb, amsmath}
\usepackage{multirow,makecell,adjustbox}
\title{CoCaRS: Correlation Calibration-Based Redundancy Suppression for Heterogeneous Knowledge Distillation}

\author{
    Fengming Yu
    Haiwei Pan\corresponding
    Kejia Zhang
    Chunling Chen
    Jian Guan
    Baoying Ma
}
\affiliations{
    \textsuperscript{\rm}College of Computer Science and Technology, Harbin Engineering University, Harbin, China\\
}

\begin{document}

\maketitle

\begin{abstract}

Knowledge distillation (KD) enables a compact student model to learn from a powerful teacher and has become an effective 
paradigm for model compression. 
The emergence of diverse model architectures has extended KD from homogeneous to heterogeneous settings.
However, differences in architectural inductive biases between the teacher and student models often result in substantial 
representation discrepancies, limiting the effectiveness of direct knowledge transfer.
Recently, redundancy suppression has offered a new perspective on heterogeneous KD by preserving cross-architecture 
invariance and reducing feature redundancy through decorrelation of teacher-student feature correlations. 
Nevertheless, this formulation may weaken useful structural information through uniform decorrelation, 
while a fixed coefficient may make the effective contribution of redundancy suppression sensitive to teacher-student 
pairs and training stages.
To address these problems, \textbf{Co}rrelation \textbf{Ca}libration-based \textbf{R}edundancy \textbf{S}uppression (\textbf{CoCaRS}) 
is proposed to better retain structural information while suppressing redundancy and reduce sensitivity to coefficient 
settings across teacher-student pairs and training stages.
Specifically, CoCaRS calibrates feature decorrelation through Confusion Evidence Estimation (CEE) and Strength Allocation 
Control (SAC), which respectively capture reliable semantic relations for correlation estimation and preserve discriminative 
structure during decorrelation. 
Adaptive Coefficient Regulation (ACR) further regulates the contribution of the calibrated redundancy suppression 
objective according to its relative loss scale, reducing sensitivity to coefficient settings.
Extensive experiments on CIFAR-100 and ImageNet-1K validate the effectiveness of CoCaRS in improving distillation 
performance and reducing sensitivity to coefficient settings.
Code will be released soon.

\end{abstract}

\section{Introduction}
Recent progress in visual recognition has been largely driven by advanced neural architectures,
such as CNNs \cite{DBLP:conf/cvpr/HeZRS16,DBLP:conf/cvpr/SandlerHZZC18,DBLP:conf/cvpr/0003MWFDX22}, 
ViTs \cite{DBLP:conf/iclr/DosovitskiyB0WZ21,DBLP:conf/icml/TouvronCDMSJ21,DBLP:conf/iccv/LiuL00W0LG21} and 
MLP-based models \cite{DBLP:conf/nips/TolstikhinHKBZU21,DBLP:journals/pami/TouvronBCCEGIJSVJ23}. 
While these models achieve strong performance, their computational and storage costs often hinder deployment in 
resource-constrained scenarios.
Knowledge distillation (KD) \cite{DBLP:journals/corr/HintonVD15} has therefore become a widely used approach for model compression, where a lightweight student
model is trained with guidance from a strong teacher to reduce inference cost with minimal performance degradation.

Existing KD methods can be categorized into response-based, feature-based, and relation-based approaches \cite{gou2021knowledge,2024-40694}. 
Response-based methods transfer probability distributions via soft targets \cite{DBLP:journals/corr/HintonVD15,DBLP:conf/aaai/YangXQY19,DBLP:conf/iccv/SonNCH21}, 
while feature-based methods use intermediate representations as additional supervision \cite{DBLP:journals/corr/RomeroBKCGB14,DBLP:conf/aaai/HeoLY019a,DBLP:conf/cvpr/LinXWYCLW22}. 
Relation-based methods further distill structural information, such as relations among samples or feature representations \cite{DBLP:conf/cvpr/YimJBK17,DBLP:conf/cvpr/ParkKLC19,DBLP:conf/iclr/TianKI20}. 

These distillation methods have demonstrated their effectiveness in homogeneous settings. 
Nevertheless, restricting distillation to such homogeneous scenarios limits the flexibility of teacher selection, 
since a high-performing teacher with the same architecture as the student may not always be available in practice. 
For heterogeneous model pairs, differences in architectural inductive biases can lead to substantial discrepancies 
between teacher and student representations, reducing the compatibility of transferred knowledge and resulting in 
mismatched supervision and suboptimal performance \cite{DBLP:conf/nips/RaghuUKZD21,DBLP:conf/accv/LiuCLHDL22,DBLP:conf/nips/HaoG0TH0X23}.

Recent heterogeneous KD studies have explored different ways to adapt teacher supervision across architectures. 
For example, OFA \cite{DBLP:conf/nips/HaoG0TH0X23} projects intermediate representations into the logit space, 
FBT \cite{DBLP:conf/iccv/LiWYDGX25} fuses heterogeneous features through an auxiliary model, 
and PAT \cite{DBLP:conf/iccv/LinYHXSC25} adapts teacher representations through feature prompting and region-aware attention. 
Different from these adaptation strategies, RSD \cite{DBLP:conf/iccv/ZhangLRM25} formulates heterogeneous KD from the perspective of redundancy suppression. 
It constructs a feature correlation matrix between teacher and student representations, where the diagonal entries 
encourage invariance across architectures, while the off-diagonal entries are constrained by a feature decorrelation 
objective to suppress redundancy, with a fixed coefficient controlling the contribution of the resulting RSD term. 
However, RSD applies a uniform decorrelation constraint to feature correlations, 
even though some of them may also encode useful structural information. 
Consequently, such information may be weakened together with redundancy. 
Moreover, as the RSD term varies in scale relative to the task loss across teacher-student pairs and training stages, 
a fixed coefficient may yield varying effective contributions and make distillation performance sensitive to coefficient selection.

To resolve these problems, \textbf{Co}rrelation \textbf{Ca}libration-based \textbf{R}edundancy \textbf{S}uppression (\textbf{CoCaRS}) 
is proposed to calibrate feature decorrelation for better preservation of structural information encoded in feature correlations, 
while adaptively regulating the effective contribution of redundancy suppression to reduce sensitivity to coefficient selection.
Specifically, CoCaRS introduces Semantic Correlation Calibration (SCC) to retain cross-architecture invariance 
while calibrating feature decorrelation through Confusion Evidence Estimation (CEE) and Strength Allocation Control (SAC). 
CEE derives confusion weights from teacher responses to capture reliable semantic relations for correlation estimation, 
whereas SAC constructs a semantic strength map from a discriminative subspace induced by the teacher classifier to 
preserve discriminative structure during decorrelation.
Adaptive Coefficient Regulation (ACR) further regulates the contribution of SCC according to its loss scale relative to 
the task objective, thereby reducing sensitivity to coefficient settings.
The main contributions of this work are summarized as follows:
\begin{itemize}
\item CoCaRS is proposed to refine redundancy suppression for heterogeneous KD through correlation calibration, retaining structural information while adaptively regulating its effective contribution.
\item In SCC, CEE captures reliable semantic relations for correlation estimation, while SAC preserves discriminative structure during decorrelation.
\item ACR adaptively regulates the effective contribution of SCC to reduce sensitivity to coefficient settings.
\item Extensive experiments on CIFAR-100 and ImageNet-1K demonstrate the effectiveness of CoCaRS across diverse heterogeneous teacher-student pairs.
\end{itemize}

\section{Related Work}
\subsection{Knowledge Distillation}
Knowledge distillation transfer knowledge from a teacher model to a smaller student model through soft labels~\cite{DBLP:journals/corr/HintonVD15}. 
Subsequent methods improve student learning with richer teacher supervision \cite{DBLP:conf/aaai/YangXQY19,DBLP:conf/iccv/SonNCH21,DBLP:conf/cvpr/LinXWYCLW22,DBLP:conf/iccv/LaoSLLY23,DBLP:conf/iclr/TianKI20,DBLP:conf/iclr/XuFZXWDX022}.

Knowledge distillation across heterogeneous architectures has also been explored. \cite{DBLP:conf/icml/TouvronCDMSJ21,DBLP:conf/cvpr/RenGHXTHZ22,DBLP:conf/accv/LiuCLHDL22,DBLP:conf/iccv/ZhaoSL23}.
These methods usually target specific architecture pairs or fixed transfer directions, limiting their applicability to 
diverse heterogeneous pairs.
Recent studies therefore explore general frameworks for heterogeneous KD.
OFA \cite{DBLP:conf/nips/HaoG0TH0X23} projects intermediate representations into logits to alleviate the semantic 
mismatch.
FBT \cite{DBLP:conf/iccv/LiWYDGX25} integrates teacher and student representations through an auxiliary model.
PAT \cite{DBLP:conf/iccv/LinYHXSC25} adapts teacher representations through prompt tuning and aligns student features through region-aware attention.
In contrast, RSD \cite{DBLP:conf/iccv/ZhangLRM25} formulates this problem from the perspective of redundancy suppression, 
preserving architecture invariant knowledge while suppressing redundant correlations. 
Our work follows this direction and further calibrates the decorrelation process.
 
\subsection{Semantic-Aware Redundancy Suppression}
Redundancy suppression has been widely studied in representation learning. 
Barlow Twins \cite{DBLP:conf/icml/ZbontarJMLD21} reduces feature redundancy by driving the cross-correlation matrix 
between two augmented views toward the identity matrix, 
while VICReg \cite{DBLP:conf/iclr/BardesPL22} regularizes covariance to reduce dependencies among embedding variables.
RSD extends this principle to heterogeneous KD through a teacher-student correlation matrix, 
whose diagonal terms preserve cross-architecture invariance and off-diagonal terms suppress redundancy through decorrelation.

Relational structures in teacher representations have also been explored in KD.
SPKD \cite{DBLP:conf/iccv/TungM19} preserves pairwise similarities between samples, 
CC \cite{DBLP:conf/iccv/PengJLZWLZ019} transfers instance correlations, 
RKD \cite{DBLP:conf/cvpr/ParkKLC19} distills distance and angle relations, 
and ICKD \cite{DBLP:conf/iccv/LiuHLXWCL21} matches inter-channel correlations. 
These methods show that correlations in teacher representations can encode structural information, 
which is relevant to the decorrelation term in redundancy suppression.

Beyond these relational structures, 
SemCKD \cite{DBLP:conf/aaai/ChenMZWWF021} addresses teacher-student semantic mismatch through adaptive cross-layer calibration.
SimKD \cite{DBLP:conf/cvpr/ChenMZWF022} reuses the teacher classifier to guide student feature learning. 
Neural Collapse \cite{DBLP:journals/corr/abs-2008-08186} reveals the geometric alignment between classifier weights and class means, 
and NCKD \cite{DBLP:conf/aaai/ZhangS025} exploits such class geometry for KD.
These works motivate semantic calibration and classifier structure in redundancy suppression.

\section{Method}
\subsection{Preliminaries}
Redundancy Suppression Distillation (RSD) introduces a redundancy suppression perspective for heterogeneous distillation.
Given teacher representations $\mathbf{f}^{t}\in\mathbb{R}^{B\times D}$ and student representations $\mathbf{f}^{s}_{ori}\in\mathbb{R}^{B\times d_s}$,
an adaptor $h(\cdot)$ maps the student representations into the teacher feature space,
yielding $\mathbf{f}^{s}\in\mathbb{R}^{B\times D}$.
Here, $B$ is the batch size, while $D$ and $d_s$ are the teacher and student feature dimensions.
Based on $\mathbf{f}^t$ and $\mathbf{f}^s$, RSD constructs a Pearson correlation matrix $\mathbf{P}\in\mathbb{R}^{D\times D}$ between 
teacher and student feature units:
\begin{equation}
\mathbf{P}_{ij} = \frac{
    \sum_{k=1}^{B}(f^t_{ki} - \bar{f}^t_{i})(f^s_{kj} - \bar{f}^s_{j})
}{
    \sqrt{\sum_{k=1}^{B}(f^t_{ki} - \bar{f}^t_{i})^2 \sum_{k=1}^{B}(f^s_{kj} - \bar{f}^s_{j})^2}
},
\end{equation}
where $\bar{f}$ denotes the mini-batch mean.
Using $\mathbf{I}_{D}$ as the target, diagonal entries are driven toward one to preserve cross-architecture invariance,
whereas off-diagonal entries are driven toward zero to suppress feature redundancy.
The RSD objective can be written as an off-diagonal reweighted mean-square error:
\begin{equation}
\mathcal{L}_{\mathrm{RSD}} = \frac{1}{D^2} [\sum_{i=1}^{D}(1-\mathbf{P}_{ii})^{2} + \kappa \sum_{i\neq j} \mathbf{P}_{ij}^{2}],
\end{equation}
where $\kappa$ controls the strength of the decorrelation objective. 
With $\beta$ weighting the RSD term, the overall objective is
\begin{equation}
\mathcal{L} = \mathcal{L}_{\mathrm{CE}} + \beta\mathcal{L}_{\mathrm{RSD}}.
\end{equation}

\subsection{Overall Framework}
\begin{figure*}[!htbp]
  \centering
  \includegraphics[width=1.0\textwidth]{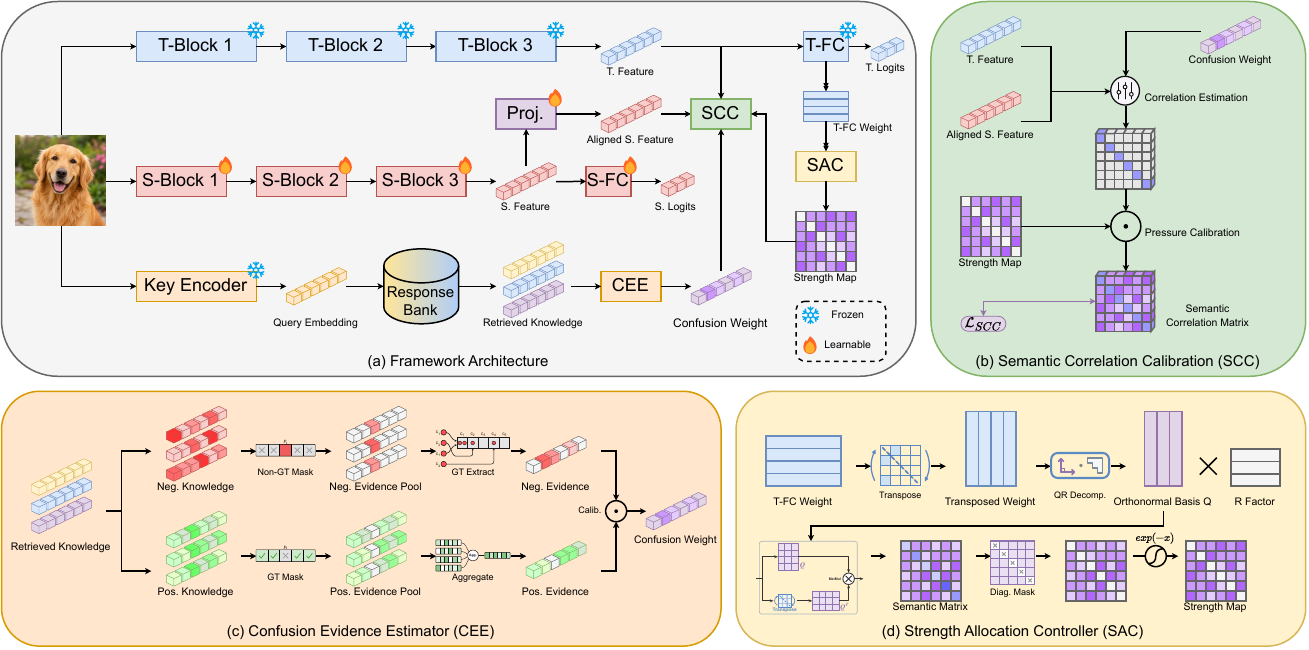} 
  \caption{Overview of the proposed CoCaRS framework. 
    (a) CoCaRS refines redundancy suppression for heterogeneous distillation through CEE and SAC;
    (b) The resulting confusion weights and strength map jointly calibrate feature decorrelation in SCC;
    (c) CEE derives confusion weights from positive and reciprocal negative evidence in retrieved knowledge;
    (d) SAC constructs the strength map from a discriminative subspace induced by the teacher classifier through QR decomposition.
}
  \label{1-Overall}
  \vspace{-0.2cm}
\end{figure*}

CoCaRS revisits redundancy suppression from the perspective of correlation calibration, as shown in Fig.~\ref{1-Overall}. 
Semantic Correlation Calibration (SCC) calibrates feature decorrelation in $\mathbf{P}$ while retaining 
cross-architecture invariance, rather than imposing a uniform constraint.
Confusion Evidence Estimation (CEE) derives confusion weights from positive and reciprocal negative evidence 
to calibrate correlation estimation, while Strength Allocation Control (SAC) constructs a semantic strength map
to allocate decorrelation strength.
Adaptive Coefficient Regulation (ACR) further regulates the SCC contribution using its loss scale relative to the task objective, 
reducing sensitivity to coefficient settings.
Overall, CoCaRS combines semantic calibration of feature decorrelation with adaptive regulation of the SCC contribution.

\subsection{Confusion Evidence Estimation}
In RSD, feature redundancy is suppressed by penalizing the off-diagonal entries of the teacher-student correlation matrix. 
CEE introduces sample awareness into correlation estimation by estimating semantic confusion evidence for each training sample.
Samples with stronger semantic confusion reflected in teacher responses are treated as more informative for correlation estimation.
In this way, sample-level importance is incorporated into correlation estimation in SCC through teacher dark knowledge 
aggregated over retrieved samples, while the original redundancy suppression objective is preserved.

Non-target teacher responses are used as semantic cues for confusion evidence, since they encode dark knowledge 
about semantic relations beyond the ground-truth label \cite{DBLP:conf/cvpr/ZhaoCSQL22}. 
Accordingly, given the training set $\mathcal{D}=\{(x_i,y_i)\}_{i=1}^{N}$ and a collection of pretrained teachers $\{T_{m}\}_{m=1}^{M}$, 
a teacher response bank $\mathcal{B}$ is constructed before distillation:
\begin{equation}
\mathcal{B} = \{
    \left(\mathbf{k}_{i}, y_{i}, \mathbf{z}^{t}_{i,1}, \ldots, \mathbf{z}^{t}_{i,M}\right)\}_{i=1}^{N},
\end{equation}
where $\mathbf{k}_{i}=\phi_k(x_i)$ is the retrieval key extracted by a pretrained key encoder $\phi_k$,
and $\mathbf{z}^{t}_{i,m}\in\mathbb{R}^{C}$ denotes the logits produced by teacher $T_m$.
For each sample $x_i$, a query embedding $\mathbf{q}_i = \phi_k(x_i)$ is used to retrieve a knowledge set $\mathcal{K}_i$ 
from $\mathcal{B}$. 
Let $\mathcal{K}_i^{+}$ denote the positive knowledge set containing retrieved samples labeled $y_i$.
Aggregating teacher logits over $\mathcal{K}_i^{+}$ provides a local estimate of the non-target responses around $x_i$.
Given the normalized teacher logits $\tilde{\mathbf{z}}_{j,m}^{t}$ for each retrieved sample $x_j$,
the positive evidence is formulated as
\begin{equation}
\mathbf{e}^{+}_{i} = \Psi_{y_i}^{+}(
    \frac{1}{M|\mathcal{K}^{+}_{i}|} \sum_{m=1}^{M}\sum_{x_j\in\mathcal{K}^{+}_i} \tilde{\mathbf{z}}_{j,m}^{t}),
\end{equation}
where $\Psi_{y_i}^{+}(\cdot)$ masks the ground-truth entry and rescales the remaining non-target entries.
The obtained $\mathbf{e}_i^{+}\in\mathbb{R}^{C}$ indicates which non-target classes are supported as potential semantic 
competitors to $y_i$ in the local neighborhood of $x_i$.

To assess whether these semantic competitors reflect credible confusion with $y_i$, negative evidence is further 
estimated from $\mathcal{K}_i^{-}$, which contains retrieved samples with labels different from $y_i$.
For each non-target class $c$, let $\mathcal{K}^{-}_{i,c} = \{x_j \in \mathcal{K}^{-}_i \mid y_j = c\}$.
The corresponding negative evidence entry evaluates the credibility of the semantic confusion between $y_i$ and $c$ 
by measuring the teacher responses to $y_i$ for samples in $\mathcal{K}^{-}_{i,c}$.
Formally, it is defined as
\begin{equation}
\tilde{e}^{-}_{i,c} = 
\left\{
    \begin{array}{lr}
            \frac{1}{M|\mathcal{K}^{-}_{i,c}|} \sum_{m=1}^{M}\sum_{x_j\in\mathcal{K}^{-}_{i,c}} \tilde{z}_{j,m,y_i}^{t}, 
            & |\mathcal{K}^{-}_{i,c}| > 0,\\
            0, &|\mathcal{K}^{-}_{i,c}| = 0.
    \end{array}
\right.
\end{equation}
The entries are assembled according to their class coordinates and calibrated as
\begin{equation}
    \mathbf{e}^{-}_{i} = \Psi^{-}(
        [\tilde{e}^{-}_{i,1}, \tilde{e}^{-}_{i,2}, \ldots, \tilde{e}^{-}_{i,C}]
    ),          
\end{equation}
where $\Psi^{-}(\cdot)$ denotes the calibration of the assembled negative evidence vector. 
The obtained $\mathbf{e}^{-}_{i} \in \mathbb{R}^C$ serves as calibrated reciprocal evidence for assessing the 
credibility of the semantic confusion indicated by $\mathbf{e}^{+}_{i}$.

Given the positive and negative evidence defined above, the final confusion weight is obtained under an asymmetric evidence model. 
The positive evidence determines the support of possible non-target confusion, whereas the negative evidence calibrates 
the credibility of this support. 
This design reflects the assumption that reciprocal responses should reinforce an existing confusion pattern rather than 
create an independent supervision signal. Accordingly, the calibrated confusion weight is defined as
\begin{equation}
\mathbf{w}^{conf}_{i} = \mathbf{e}^{+}_{i} \odot(1+\gamma \mathbf{e}^{-}_{i}),
\end{equation}
where $\gamma$ controls the strength of reciprocal calibration. 
The resulting $\mathbf{w}^{conf}_i \in \mathbb{R}^C$ 
is a calibrated confusion weight vector that captures reliable semantic relations between $y_i$ and 
its non-target competitors for subsequent correlation estimation.

\subsection{Strength Allocation Control}
SAC calibrates the off-diagonal decorrelation term through a semantic strength map that allocates decorrelation 
strength according to associations between feature dimensions.
Neural collapse theory relates classifier weight vectors to class-level feature prototypes at convergence \cite{DBLP:journals/corr/abs-2008-08186}, 
and NCKD \cite{DBLP:conf/aaai/ZhangS025} further exploits this theory for classifier construction in distillation. 
Motivated by this, SAC uses the teacher classifier weights to induce a discriminative subspace for constructing the 
semantic strength map.

Given the normalized teacher classifier weights $\tilde{\mathbf{W}}$, an orthonormal basis $\mathbf{Q}$ for the 
discriminative subspace of the teacher classifier is obtained from $\tilde{\mathbf{W}}^{\top}$ through reduced QR decomposition.
When $\tilde{\mathbf{W}}^{\top}$ has full row rank, rank reduction is applied before QR decomposition. 
The target rank is estimated from the stable rank, which measures the effective rank of a matrix.
The semantic matrix is therefore defined as $\mathbf{M}^{sem} = |\mathbf{QQ}^{\top}|$, 
which captures the projection structure of the induced discriminative subspace over feature dimensions.
Since the diagonal entries serve as invariance anchors rather than decorrelation terms, 
they are excluded to obtain the off-diagonal component $\mathbf{M}^{off} = \mathbf{M}^{sem} \odot (1 - \mathbf{I})$. 
The strength map is defined as
\begin{equation}
    \mathbf{M}^{\kappa}_{ij} = \frac{
        \exp(-\tau_\kappa M^{off}_{ij})}{
        \frac{1}{D(D-1)}\sum_{a\neq b} \exp(-\tau_\kappa M^{off}_{ab})}, i \neq j,
\end{equation}
where $\tau_\kappa$ controls the semantic modulation strength. 
A larger $\mathbf{M}^{off}_{ij}$ indicates a stronger association between the corresponding feature dimensions within 
the induced discriminative subspace, thereby reducing the decorrelation strength applied to the corresponding feature
correlation to preserve discriminative structure.

\subsection{Distillation Formulation}
Given the confusion weights and the semantic strength map defined above, 
SCC integrates them into the decorrelation objective while preserving the diagonal invariance term.
Let $\hat{\mathbf{f}}^{t}$ and $\hat{\mathbf{f}}^{s}$ denote the normalized teacher and student features. 
The SCC objective is formulated as
\begin{equation}
    \begin{aligned}
    \mathcal{L}_{\mathrm{SCC}} &= \frac{1}{D^2} \Bigg[\sum_{i=1}^{D} (1-\mathbf{P}_{ii})^{2} \\ 
    & + \kappa\sum_{i\neq j}\mathbf{M}^{\kappa}_{ij}\left(
        \sum_{k=1}^{B} \mathcal{S}_{k}^{conf} \hat{f}^{t}_{k,i}\hat{f}^{s}_{k,j}\right)^{2}
    \Bigg].
    \end{aligned}
\end{equation}
In this objective, $\kappa$ controls the overall decorrelation strength, while the strength map $\mathbf{M}^{\kappa}$ specifies 
its relative coefficients. $\mathcal{S}_{k}^{conf}$ is derived from the confusion weight as
\begin{equation}
    \mathcal{S}_{k}^{conf} = \mathrm{Norm} \left(1+\alpha \|\mathbf{w}^{conf}_k\|_{1}\right),
\end{equation}
where $\alpha$ controls the effect of confusion evidence on sample weighting in correlation estimation.

A basic training objective combines the cross-entropy task loss with the SCC term as
\begin{equation}
\mathcal{L} = \mathcal{L}_{\mathrm{CE}} + \lambda\mathcal{L}_{\mathrm{SCC}},
\end{equation}
where $\lambda$ controls the strength of the SCC term.
However, the relative scale of the SCC term can vary across training stages and teacher-student pairs, 
making a static coefficient less suitable for maintaining a balanced optimization process. 
To address this, Adaptive Coefficient Regulation (ACR) regulates the effective contribution of SCC according to 
its relative loss scale with respect to the task objective:
\begin{equation}
r_t = \frac{\mathcal{L}_{\mathrm{SCC}}^{t}}{\mathcal{L}_{\mathrm{CE}}^{t}+\epsilon}.
\end{equation}
The SCC coefficient is regulated according to the deviation of $r_t$ from the target ratio $\rho$.
To reduce fluctuations in loss magnitudes, the coefficient is updated through EMA as
\begin{equation}
\lambda_t = \eta\lambda_{t-1} + (1-\eta) \mathcal{G}\left(\frac{\rho}{r_t+\epsilon}\right),
\end{equation}
where $\eta$ denotes the EMA decay factor, and $\mathcal{G}(\cdot)$ is a bounded modulation function.
The final objective is defined as
\begin{equation}
\mathcal{L} = \mathcal{L}_{\mathrm{CE}} + \lambda_t\mathcal{L}_{\mathrm{SCC}},
\end{equation}

%%%%%%%%%%%%%%%%%%%%%%%%%%%%%%%%%%%%%%%%%%%%%%%%%%%%%%%%%%%%%%%%%%%%%%%%%%%%%%%%%%%%%%%%%%%%%%%%%%%%%%%%%%%%%%%%%%%%%%%%
%%% Table 1. Main results on CIFAR-100
%%%%%%%%%%%%%%%%%%%%%%%%%%%%%%%%%%%%%%%%%%%%%%%%%%%%%%%%%%%%%%%%%%%%%%%%%%%%%%%%%%%%%%%%%%%%%%%%%%%%%%%%%%%%%%%%%%%%%%%%
\begin{table*}[!htbp]
\centering
\footnotesize
\caption{Top-1 accuracy (\%) on CIFAR-100.
The best and second best results are in \textbf{bold} and \underline{underlined}.}
\label{tab:cifar100}
\begin{adjustbox}{max width=\textwidth}
\begin{tabular}{@{}c|cc|cc|cccc|ccc|cccc@{}}
\toprule
\multirow{2}{*}{} & \multirow{2}{*}{Teacher} & \multirow{2}{*}{Student}
& \multicolumn{2}{c|}{From Scratch}
& \multicolumn{4}{c|}{Feature-based}
& \multicolumn{3}{c|}{Response-based}
& \multicolumn{4}{c}{Heterogeneous-KD} \\
\cmidrule(r){4-5} \cmidrule(r){6-9} \cmidrule(r){10-12} \cmidrule(r){13-16}
~ & ~ & ~ & T. & S. & FitNet & CC & RKD & CRD & KD & DKD & DIST & OFA & PAT & RSD & CoCaRS \\
\midrule

\multirow{6}{*}{\makecell{CNN-based\\students}}
 & Swin-T     & ResNet18    & 89.26 & 74.01 & 78.87 & 74.19 & 74.11 & 77.63 & 78.74 & 80.26 & 77.75 & 80.54 & 81.22 & \underline{83.92} & \textbf{85.42} \\
 & ViT-S      & ResNet18    & 92.04 & 74.01 & 77.71 & 74.26 & 73.72 & 76.60 & 77.26 & 78.10 & 76.49 & 80.15 & 80.11 & \underline{81.50} & \textbf{85.22} \\
 & Mixer-B/16 & ResNet18    & 87.29 & 74.01 & 77.15 & 74.26 & 73.75 & 76.42 & 77.79 & 78.67 & 76.36 & 79.39 & 80.07 & \underline{81.85} & \textbf{83.85} \\
 & Swin-T     & MobileNetV2 & 89.26 & 73.68 & 74.28 & 71.19 & 69.00 & 79.80 & 74.68 & 71.07 & 72.89 & 80.98 & 78.78 & \underline{83.68} & \textbf{85.50} \\
 & ViT-S      & MobileNetV2 & 92.04 & 73.68 & 73.54 & 70.67 & 68.46 & 78.14 & 72.77 & 69.80 & 72.54 & 78.45 & 78.87 & \underline{81.68} & \textbf{85.62} \\
 & Mixer-B/16 & MobileNetV2 & 87.29 & 73.68 & 73.78 & 70.73 & 68.95 & 78.15 & 73.33 & 70.20 & 73.26 & 78.78 & 78.62 & \underline{81.74} & \textbf{84.46} \\
\midrule

\multirow{4}{*}{\makecell{ViT-based\\students}}
 & ConvNeXt-T & DeiT-T      & 88.41 & 68.00 & 60.78 & 68.01 & 69.79 & 65.94 & 72.99 & 74.60 & 73.55 & 75.76 & 79.59 & \underline{82.46} & \textbf{84.09} \\
 & Mixer-B/16 & DeiT-T      & 87.29 & 68.00 & 71.05 & 68.13 & 69.89 & 65.35 & 71.36 & 73.44 & 71.67 & 73.90 & 74.66 & \underline{78.50} & \textbf{81.54} \\
 & ConvNeXt-T & Swin-P      & 88.41 & 72.63 & 24.06 & 72.63 & 71.73 & 67.09 & 76.44 & 76.80 & 76.41 & 78.32 & 80.74 & \underline{82.21} & \textbf{85.08} \\
 & Mixer-B/16 & Swin-P      & 87.29 & 72.63 & 75.20 & 73.32 & 70.82 & 67.03 & 75.93 & 76.39 & 75.85 & 76.65 & 78.44 & \underline{81.28} & \textbf{84.05} \\
\midrule

\multirow{2}{*}{\makecell{MLP-based\\students}}
 & ConvNeXt-T & ResMLP-S12  & 88.41 & 66.56 & 45.47 & 67.70 & 65.82 & 63.35 & 72.25 & 73.22 & 71.93 & 75.21 & 83.50 & \underline{84.21} & \textbf{86.63} \\
 & Swin-T     & ResMLP-S12  & 89.26 & 66.56 & 63.12 & 68.37 & 64.66 & 61.72 & 71.89 & 72.82 & 11.05 & 73.58 & 80.94 & \underline{82.67} & \textbf{84.99} \\
\midrule

\multicolumn{3}{c|}{Average}
 & 88.85 & 71.45 & 66.25 & 71.12 & 70.06 & 71.44& 74.62 & 74.61 & 69.15& 77.64 & 79.63 & \underline{82.14} & \textbf{84.70} \\
\bottomrule
\end{tabular}
\end{adjustbox}
\vspace{-0.2cm}
\end{table*}

\section{Experiments}
\subsection{Experimental Setup}
\subsubsection{Models}
Teacher-student pairs are constructed from models with different architectures. 
For CNN-based models, ResNet \cite{DBLP:conf/cvpr/HeZRS16}, MobileNetV2 \cite{DBLP:conf/cvpr/SandlerHZZC18}, and ConvNeXt \cite{DBLP:conf/cvpr/0003MWFDX22} are adopted. 
Transformer-based models cover ViT \cite{DBLP:conf/iclr/DosovitskiyB0WZ21}, DeiT \cite{DBLP:conf/icml/TouvronCDMSJ21}, Swin Transformer \cite{DBLP:conf/iccv/LiuL00W0LG21}, and its lightweight variants, Swin-Pico and Swin-Nano \cite{DBLP:conf/nips/HaoG0TH0X23}. 
MLP-based models include MLP-Mixer \cite{DBLP:conf/nips/TolstikhinHKBZU21} and ResMLP \cite{DBLP:journals/pami/TouvronBCCEGIJSVJ23}.

\subsubsection{Datasets}
Experiments are conducted on CIFAR-100 \cite{krizhevsky2009learning} and ImageNet-1K \cite{deng2009imagenet}. 
CIFAR-100 contains 60,000 images from 100 classes, with 50,000 images used for training and 10,000 images used for testing. 
ImageNet-1K is a large-scale dataset containing approximately 1.28 million training images and 50,000 validation images 
from 1,000 classes.

\subsubsection{Baselines}
Several representative KD methods are selected for comparison. 
Feature-based methods include FitNet \cite{DBLP:journals/corr/RomeroBKCGB14}, CC \cite{DBLP:conf/iccv/PengJLZWLZ019}, RKD \cite{DBLP:conf/cvpr/ParkKLC19}, and CRD \cite{DBLP:conf/iclr/TianKI20}, 
which transfer knowledge through intermediate representations or feature relations. 
Response-based methods include KD \cite{DBLP:journals/corr/HintonVD15}, DKD \cite{DBLP:conf/cvpr/ZhaoCSQL22}, and DIST \cite{DBLP:conf/nips/0020Y00022}, 
which align the output responses of teacher and student models. 
Heterogeneous KD methods, including OFA \cite{DBLP:conf/nips/HaoG0TH0X23}, PAT \cite{DBLP:conf/iccv/LinYHXSC25}, and RSD \cite{DBLP:conf/iccv/ZhangLRM25}, are also compared.

\subsection{Main Results}

\subsubsection{Results on CIFAR-100}
Experiments are conducted on CIFAR-100 using heterogeneous teacher-student pairs involving CNNs, ViTs, and MLPs. 
As reported in Table~\ref{tab:cifar100}, CoCaRS achieves the best results across all evaluated pairs, with an average accuracy of 84.70\%.
Feature-based and response-based methods obtain average Top-1 accuracies of 69.72\% and 72.79\%, respectively, 
and the former is below the scratch baseline of 71.45\%.
These results indicate the limited applicability of homogeneous methods to heterogeneous teacher-student pairs, 
whereas methods designed for heterogeneous KD achieve higher average accuracies.
RSD achieves an average accuracy of 82.14\% through redundancy suppression, 
while CoCaRS raises the average accuracy to 84.70\% by calibrating feature decorrelation, yielding a 2.56\% gain.
This result supports the benefit of calibrating feature decorrelation rather than applying a uniform constraint.

\subsubsection{Results on ImageNet-1K}
CoCaRS further achieves the highest average Top-1 accuracy of 74.46\% on ImageNet-1K, 
outperforming RSD by 0.43\% points, as reported in Table~\ref{tab:imagenet}. 
Feature-based and response-based methods obtain lower average accuracies than methods developed for heterogeneous KD, 
reflecting the difficulty of transferring knowledge across heterogeneous architectures. 
The performance of RSD demonstrates the effectiveness of redundancy suppression in this setting, 
while the further improvement achieved by CoCaRS supports the benefit of correlation calibration beyond the original 
formulation. 
Notably, CoCaRS yields its largest improvement over RSD on Swin-T $\rightarrow$ ResMLP-S12, with a gain of 0.85\%.
These results further validate correlation calibration for large-scale heterogeneous distillation.

%%%%%%%%%%%%%%%%%%%%%%%%%%%%%%%%%%%%%%%%%%%%%%%%%%%%%%%%%%%%%%%%%%%%%%%%%%%%%%%%%%%%%%%%%%%%%%%%%%%%%%%%%%%%%%%%%%%%%%%%
%%% Table 2. Main results on ImageNet-1K
%%%%%%%%%%%%%%%%%%%%%%%%%%%%%%%%%%%%%%%%%%%%%%%%%%%%%%%%%%%%%%%%%%%%%%%%%%%%%%%%%%%%%%%%%%%%%%%%%%%%%%%%%%%%%%%%%%%%%%%%
\begin{table*}[!htbp]
\centering
\footnotesize
\caption{Top-1 accuracy (\%) on ImageNet-1K.
The best and second best results are in \textbf{bold} and \underline{underlined}.}
\label{tab:imagenet}
\begin{adjustbox}{max width=\textwidth}
\begin{tabular}{@{}cc|cc|cccc|ccc|cccc@{}}
\toprule
\multirow{2}{*}{Teacher} & \multirow{2}{*}{Student}
& \multicolumn{2}{c|}{From Scratch}
& \multicolumn{4}{c|}{Feature-based}
& \multicolumn{3}{c|}{Response-based}
& \multicolumn{4}{c}{Heterogeneous-KD} \\
\cmidrule(r){3-4} \cmidrule(r){5-8} \cmidrule(r){9-11} \cmidrule(r){12-15}
~ & ~ & T. & S. & FitNet & CC & RKD & CRD & KD & DKD & DIST & OFA & PAT & RSD & CoCaRS \\
\midrule

\multicolumn{15}{l}{\textit{CNN-based students}} \\
\midrule
Swin-T & ResNet18            & 81.35 & 69.75 & 71.18 & 70.07 & 68.89 & 69.09 & 71.14 & 71.10 & 70.91 & 71.85 & 71.54          & \underline{72.13} & \textbf{72.63} \\
Mixer-B/16 & MobileNetV2     & 76.55 & 68.87 & 71.59 & 70.79 & 69.86 & 68.89 & 71.92 & 70.93 & 71.74 & 72.12 & \textbf{72.22} & 71.90             & \underline{72.15} \\
\midrule
\multicolumn{15}{l}{\textit{ViT-based students}} \\
\midrule
ConvNeXt-T & DeiT-T          & 82.05 & 72.17 & 70.45 & 73.12 & 71.47 & 69.18 & 74.00 & 73.95 & 74.07 & 74.41 & 74.44 & \underline{74.46} & \textbf{74.59} \\
\midrule
\multicolumn{15}{l}{\textit{MLP-based students}} \\
\midrule
Swin-T & ResMLP-S12          & 81.35 & 76.65 & 76.48 & 76.15 & 75.10 & 73.40 & 76.67 & 76.99 & 77.25 & 77.31 & 77.59 & \underline{77.61} & \textbf{78.46} \\
\midrule
\multicolumn{2}{c|}{Average} & 80.33 & 71.86 & 72.43 & 72.53 & 71.33 & 70.14 & 73.43 & 73.24 & 73.49 & 73.92 & 73.95 & \underline{74.03} & \textbf{74.46} \\
\bottomrule
\end{tabular}
\end{adjustbox}
\vspace{-0.2cm}
\end{table*}

\subsection{Ablation Study}
\subsubsection{Effect of Core Components}
Table~\ref{tab:ablation-main} shows that removing CEE, SAC, or ACR consistently degrades performance.
The performance drops caused by removing CEE or SAC support calibrating both correlation estimation and decorrelation 
strength rather than applying uniform decorrelation.
Meanwhile, the degradation caused by removing ACR supports the need to regulate the effective contribution of SCC according to 
its loss scale relative to the task objective.
Together, these results validate semantic calibration in SCC and adaptive regulation of its contribution.

%%%%%%%%%%%%%%%%%%%%%%%%%%%%%%%%%%%%%%%%%%%%%%%%%%%%%%%%%%%%%%%%%%%%%%%%%%%%%%%%%%%%%%%%%%%%%%%%%%%%%%%%%%%%%%%%%%%%%%%%
%%% Table 3. Effect of Core Components
%%%%%%%%%%%%%%%%%%%%%%%%%%%%%%%%%%%%%%%%%%%%%%%%%%%%%%%%%%%%%%%%%%%%%%%%%%%%%%%%%%%%%%%%%%%%%%%%%%%%%%%%%%%%%%%%%%%%%%%%
\begin{table}[htbp]
\centering
\scriptsize
\setlength{\tabcolsep}{1.5pt}
\caption{Ablation study on CIFAR-100: effect of the core components of CoCaRS.}
\label{tab:ablation-main}
\begin{adjustbox}{max width=\columnwidth}
\begin{tabular}{@{}ccc|cccccc@{}}
\toprule
\makecell{w/\\CEE} & \makecell{w/\\SAC} & \makecell{w/\\ACR} & \makecell{Swin-T\\ResNet18} & \makecell{Mixer-B/16\\ResNet18} & \makecell{ViT-S\\MobileNetV2} & \makecell{ConvNeXt-T\\DeiT-T} & \makecell{Mixer-B/16\\Swin-P} & \makecell{ConvNeXt-T\\ResMLP-S12} \\
\midrule
\multicolumn{3}{c|}{RSD Baseline}    & 83.92             & 81.85             & 81.68             & 82.46             & 81.28             & 84.21 \\
\midrule
\checkmark & --         & --         & 84.38             & 82.45             & 83.72             & 82.95             & 83.04             & 85.20 \\
--         & \checkmark & --         & 84.52             & 82.60             & 83.96             & 82.93             & 83.35             & 85.50 \\
\midrule
--         & \checkmark & \checkmark & \underline{84.81} & 83.60             & \underline{85.09} & 83.54             & 83.50             & 85.69 \\
\checkmark & --         & \checkmark & 84.71             & 83.37             & 84.94             & \underline{83.65} & 83.44             & 85.53 \\
\checkmark & \checkmark & --         & 84.50             & \underline{83.62} & 84.91             & 83.62             & \underline{83.65} & \underline{85.94} \\
\midrule
\checkmark & \checkmark & \checkmark & \textbf{85.42}    & \textbf{83.85}    & \textbf{85.62}    & \textbf{84.09}    & \textbf{84.05}    & \textbf{86.63} \\
\bottomrule
\end{tabular}
\end{adjustbox}
\vspace{-0.2cm}
\end{table}

\subsubsection{Effect of CEE Formulation}
The construction of confusion evidence from retrieved knowledge and the reciprocal calibration provided by negative 
evidence are examined in Table~\ref{tab:ablation-ce}. 
Removing CEE entirely reduces performance, confirming the contribution of confusion evidence to correlation estimation in SCC.
Removing negative evidence (\emph{w/o NE}) also leads to lower performance, 
supporting its contribution to calibrating the credibility of the semantic confusion indicated by positive evidence. 
Replacing the retrieved samples with randomly selected samples (\emph{Random}) results in a clear performance decrease, 
suggesting that an effective local estimate of the non-target responses relies on retrieved samples associated with the 
query sample. 
These results support the use of retrieved knowledge for positive evidence estimation and negative evidence for 
reciprocal calibration.

%%%%%%%%%%%%%%%%%%%%%%%%%%%%%%%%%%%%%%%%%%%%%%%%%%%%%%%%%%%%%%%%%%%%%%%%%%%%%%%%%%%%%%%%%%%%%%%%%%%%%%%%%%%%%%%%%%%%%%%%
%%% Table 4. Effect of CEE Formulation
%%%%%%%%%%%%%%%%%%%%%%%%%%%%%%%%%%%%%%%%%%%%%%%%%%%%%%%%%%%%%%%%%%%%%%%%%%%%%%%%%%%%%%%%%%%%%%%%%%%%%%%%%%%%%%%%%%%%%%%%
\begin{table}[htbp]
\centering
\scriptsize
\setlength{\tabcolsep}{2pt}
\caption{Ablation study on CIFAR-100: effects of retrieved knowledge and negative evidence in CEE.}
\label{tab:ablation-ce}
\begin{adjustbox}{max width=\columnwidth}
\begin{tabular}{@{}lcccccc@{}}
\toprule
\makecell[l]{CEE \\ Setting} & \makecell{Swin-T\\ResNet18} & \makecell{Mixer-B/16\\ResNet18} & \makecell{ViT-S\\MobileNetV2} & \makecell{ConvNeXt-T\\DeiT-T} & \makecell{Mixer-B/16\\Swin-P} & \makecell{ConvNeXt-T\\ResMLP-S12} \\
\midrule
w/o CEE   & 84.81 & 83.60 & 85.09 & 83.54 & 83.50 & 85.69 \\
w/o NE    & 84.50 & 83.60 & 85.04 & 83.35 & 83.33 & 85.76 \\
Random    & 84.60 & 83.22 & 84.64 & 83.20 & 82.55 & 86.05 \\
\midrule
CoCaRS & \textbf{85.42} & \textbf{83.85} & \textbf{85.62} & \textbf{84.09} & \textbf{84.05} & \textbf{86.63} \\
\bottomrule
\end{tabular}
\end{adjustbox}
\vspace{-0.2cm}
\end{table}

\subsubsection{Effect of SAC Formulation}
The construction of the semantic strength map from the induced discriminative subspace and the direction of 
decorrelation strength allocation are examined in Table~\ref{tab:ablation-sac}. 
Removing SAC entirely reduces performance, showing the contribution of the semantic strength map to decorrelation strength allocation.
When the allocation direction is inverted (\emph{Inverted}), larger values of $\mathbf{M}^{off}$ lead to greater 
decorrelation strength, while the resulting performance remains close to that without SAC. 
This contrast supports allocating lower decorrelation strength to stronger associations in the induced 
discriminative subspace.
Replacing the induced discriminative subspace with a random orthogonal subspace (\emph{Random}) also performs worse than 
the complete formulation, supporting the use of teacher classifier weights as the structural basis of the semantic strength map. 
These results further validate the proposed direction of decorrelation strength allocation.

%%%%%%%%%%%%%%%%%%%%%%%%%%%%%%%%%%%%%%%%%%%%%%%%%%%%%%%%%%%%%%%%%%%%%%%%%%%%%%%%%%%%%%%%%%%%%%%%%%%%%%%%%%%%%%%%%%%%%%%%
%%% Table 5. Effect of SAC Formulation
%%%%%%%%%%%%%%%%%%%%%%%%%%%%%%%%%%%%%%%%%%%%%%%%%%%%%%%%%%%%%%%%%%%%%%%%%%%%%%%%%%%%%%%%%%%%%%%%%%%%%%%%%%%%%%%%%%%%%%%%
\begin{table}[htbp]
\centering
\scriptsize
\setlength{\tabcolsep}{2pt}
\caption{Ablation study on CIFAR-100: effects of strength allocation and discriminative subspace in SAC.}
\label{tab:ablation-sac}
\begin{adjustbox}{max width=\columnwidth}
\begin{tabular}{@{}lcccccc@{}}
\toprule
\makecell[l]{SAC \\ Setting} & \makecell{Swin-T\\ResNet18} & \makecell{Mixer-B/16\\ResNet18} & \makecell{ViT-S\\MobileNetV2} & \makecell{ConvNeXt-T\\DeiT-T} & \makecell{Mixer-B/16\\Swin-P} & \makecell{ConvNeXt-T\\ResMLP-S12} \\
\midrule
w/o SAC  & 84.71 & 83.37 & 84.94 & 83.65 & 83.44 & 85.53 \\
Inverted & 84.68 & 83.31 & 85.03 & 83.24 & 83.58 & 85.73 \\
Random   & 84.96 & 83.57 & 84.90 & 83.78 & 83.33 & 86.03 \\
\midrule
CoCaRS & \textbf{85.42} & \textbf{83.85} & \textbf{85.62} & \textbf{84.09} & \textbf{84.05} & \textbf{86.63} \\
\bottomrule
\end{tabular}
\end{adjustbox}
\vspace{-0.2cm}
\end{table}

\subsubsection{Effect of Diagonal Invariance}
The role of diagonal invariance is examined in Table~\ref{tab:ablation-diagonal}. 
Extending the modulation introduced by CEE, SAC, or both to the diagonal term consistently degrades performance. 
The diagonal term is intended to preserve cross-architecture invariance between heterogeneous representations.
Applying confusion weights or the semantic strength map to this term may weaken its role in preserving cross-architecture invariance.
These results support preserving diagonal invariance while restricting CEE and SAC to off-diagonal redundancy suppression.

%%%%%%%%%%%%%%%%%%%%%%%%%%%%%%%%%%%%%%%%%%%%%%%%%%%%%%%%%%%%%%%%%%%%%%%%%%%%%%%%%%%%%%%%%%%%%%%%%%%%%%%%%%%%%%%%%%%%%%%%
%%% Table 6. Effect of Diagonal Invariance
%%%%%%%%%%%%%%%%%%%%%%%%%%%%%%%%%%%%%%%%%%%%%%%%%%%%%%%%%%%%%%%%%%%%%%%%%%%%%%%%%%%%%%%%%%%%%%%%%%%%%%%%%%%%%%%%%%%%%%%%
\begin{table}[htbp]
\centering
\scriptsize
\setlength{\tabcolsep}{1.6pt}
\caption{Ablation study on CIFAR-100: effect of CEE and SAC on the invariance term of SCC.}
\label{tab:ablation-diagonal}
\begin{adjustbox}{max width=\columnwidth}
\begin{tabular}{cc|cccccc}
\toprule
\multicolumn{2}{c|}{Invariance Term}
            & Swin-T   & Mixer-B/16 & ViT-S       & ConvNeXt-T & Mixer-B/16 & ConvNeXt-T \\
CEE & SAC   & ResNet18 & ResNet18   & MobileNetV2 & DeiT-T     & Swin-P     & ResMLP-S12 \\
\midrule
\checkmark & --         & 84.48 & 83.46 & 84.99 & 83.01 & 82.95 & 85.28 \\
--         & \checkmark & 85.02 & 83.51 & 84.78 & 83.19 & 83.41 & 85.81 \\
\checkmark & \checkmark & 84.81 & 83.11 & 84.60 & 82.60 & 83.40 & 85.94 \\
\midrule
--         & --         & \textbf{85.42} & \textbf{83.85} & \textbf{85.62} & \textbf{84.09} & \textbf{84.05} & \textbf{86.63} \\
\bottomrule
\end{tabular}
\end{adjustbox}
\vspace{-0.2cm}
\end{table}

%%%%%%%%%%%%%%%%%%%%%%%%%%%%%%%%%%%%%%%%%%%%%%%%%%%%%%%%%%%%%%%%%%%%%%%%%%%%%%%%%%%%%%%%%%%%%%%%%%%%%%%%%%%%%%%%%%%%%%%%
%%% Table 7. Results on Homogeneous Distillation
%%%%%%%%%%%%%%%%%%%%%%%%%%%%%%%%%%%%%%%%%%%%%%%%%%%%%%%%%%%%%%%%%%%%%%%%%%%%%%%%%%%%%%%%%%%%%%%%%%%%%%%%%%%%%%%%%%%%%%%%
\begin{table*}[ht]
\centering
\footnotesize
\caption{Top-1 accuracy (\%) on ImageNet-1K for homo. settings.
The best and second best results are in \textbf{bold} and \underline{underlined}.}
\label{tab:homogeneous-imagenet}
\begin{adjustbox}{max width=\textwidth}
\begin{tabular}{@{}cc|cc|cccccccccc|ccc@{}}
\toprule
\multirow{2}{*}{Teacher}
& \multirow{2}{*}{Student}
& \multicolumn{2}{c|}{From Scratch}
& \multicolumn{10}{c|}{Homogeneous-KD}
& \multicolumn{3}{c}{Heterogeneous-KD} \\
\cmidrule(r){3-4}
\cmidrule(r){5-14}
\cmidrule(r){15-17}
~ & ~ & T. & S. & KD & OFD & CRD & RKD & CAT & SimKD & Review & DKD & SDD & DIST & OFA & RSD & CoCaRS \\
\midrule
ResNet34 & ResNet18  & 73.31 & 69.75 & 70.66 & 70.81 & 71.17 & 71.34 & 71.26 & 71.59 & 71.61 & 71.70 & 71.14 & 72.07 & 72.10             & \underline{72.18} & \textbf{72.58} \\
ResNet50 & MobileNet & 80.36 & 68.58 & 68.58 & 71.25 & 71.37 & 71.32 & 72.24 & 72.25 & 72.56 & 72.05 & 72.24 & 73.24 & \underline{73.28} & 73.08             & \textbf{74.91} \\
\bottomrule
\end{tabular}
\end{adjustbox}
\vspace{-0.2cm}
\end{table*}

\subsection{Further Analysis}
\subsubsection{Performance in Homogeneous Settings}
To further evaluate CoCaRS under homogeneous settings, experiments are conducted on two teacher-student pairs on ImageNet-1K. 
The comparison additionally includes OFD \cite{DBLP:conf/iccv/HeoKYPK019}, CAT \cite{DBLP:conf/cvpr/GuoYL023}, SimKD \cite{DBLP:conf/cvpr/ChenMZWF022}, Review \cite{DBLP:conf/cvpr/Chen0ZJ21}, and SDD \cite{DBLP:conf/cvpr/WeiLL24}. 
CoCaRS achieves the best performance and outperforms RSD on both pairs, as reported in Table~\ref{tab:homogeneous-imagenet}.
These results further support the effectiveness of introducing correlation calibration into redundancy suppression. 
Together with the results on heterogeneous teacher-student pairs, this additional evaluation demonstrates the robustness 
of CoCaRS across different distillation settings.

\subsubsection{Intermediate Representation Similarity}
\begin{figure}[htbp]
  \centering
  \includegraphics[width=1.0\columnwidth]{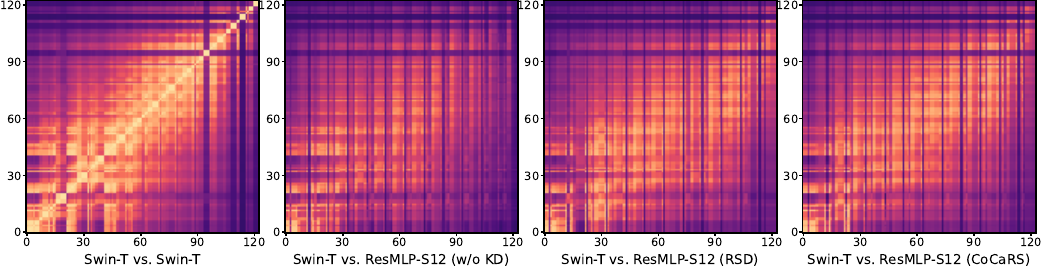} 
    \caption{Intermediate representation similarity between Swin-T and ResMLP-S12 measured by CKA, 
    where brighter regions indicate higher similarity.}  
    \label{fig:cka}
    \vspace{-0.2cm}
\end{figure}

The similarity between the intermediate representations of a Swin-T teacher and
a ResMLP-S12 student is visualized using CKA \cite{pmlr-v97-kornblith19a} in Fig.~\ref{fig:cka}.
Without KD, the student exhibits relatively low feature similarity with the teacher,
reflecting the representation discrepancy between heterogeneous architectures.
Both RSD and CoCaRS improve the similarity between teacher and student intermediate
features. Compared with RSD, CoCaRS exhibits higher similarity, particularly in the
shallow and deep regions.
The reduced representation discrepancy is consistent with the improved distillation
performance and further supports the effectiveness of CoCaRS.

\subsubsection{Computational Cost}
\begin{figure}[htbp]
  \centering
  \includegraphics[width=1.0\columnwidth]{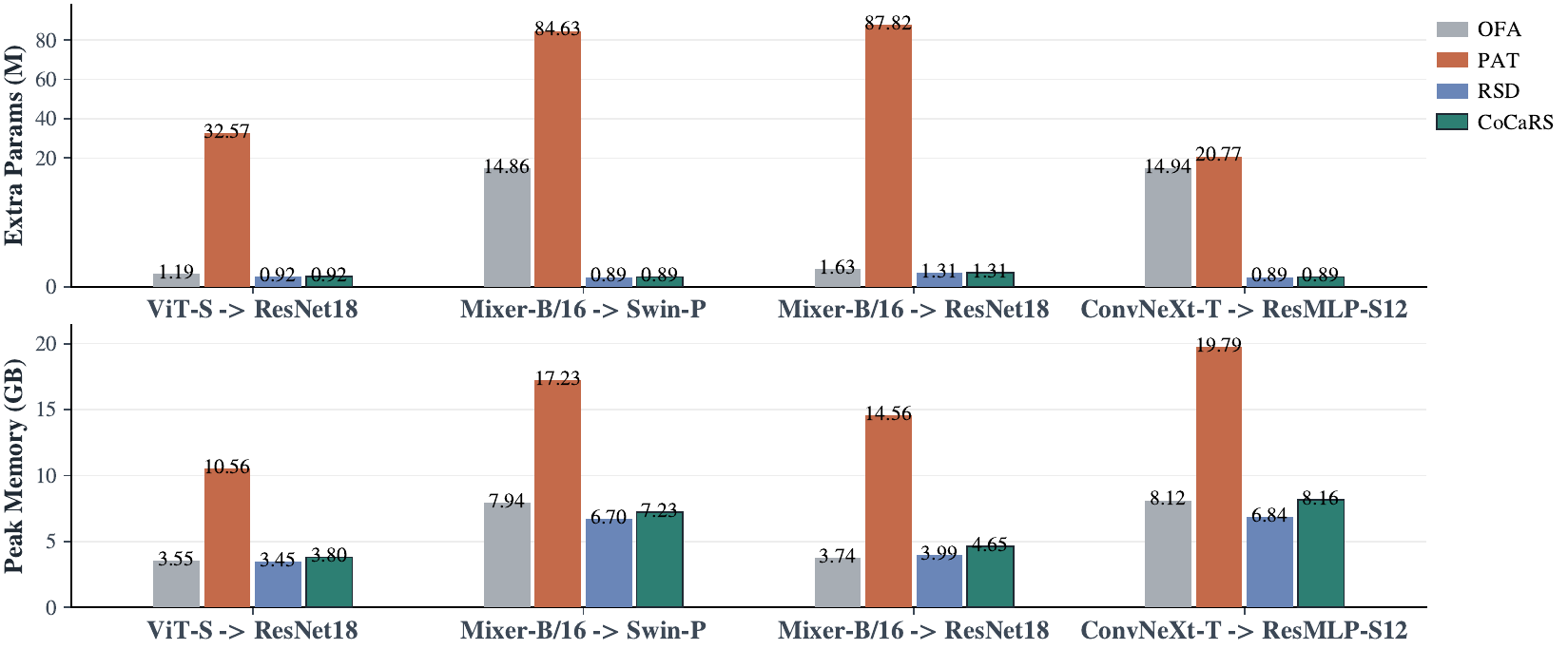} 
    \caption{Comparison of additional trainable parameters and peak memory usage on CIFAR-100.}  
    \label{fig:com-cost}
    \vspace{-0.2cm}
\end{figure}

A comparison of additional trainable parameters and peak memory usage across methods is presented in Fig.~\ref{fig:com-cost}.
CoCaRS requires fewer additional trainable parameters than OFA and PAT while matching that of RSD, 
since its proposed components introduce no additional learnable parameters. 
Its peak memory usage is comparable to that of OFA and substantially lower than that of PAT. 
Compared to RSD, CoCaRS achieves a better overall balance between distillation performance and training memory, 
despite its moderate increase in memory overhead. 
Since the additional components are used only during distillation, the original inference cost of the student is preserved.

\subsubsection{ACR for SCC Balance}
\begin{figure}[htbp]
  \centering
  \includegraphics[width=1.0\columnwidth]{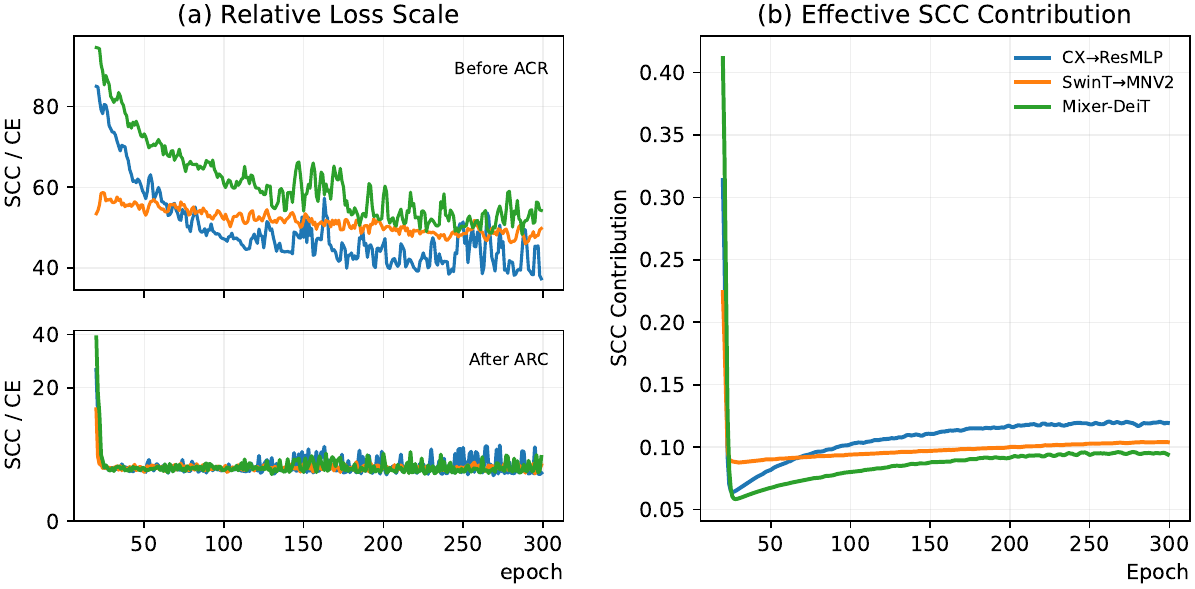} 
  \caption{ACR regulates the effective contribution of SCC across heterogeneous model pairs during training.}
  \label{fig:acr}
  \vspace{-0.2cm}
\end{figure}

As shown in Fig.~\ref{fig:acr}(a), 
the relative scale of SCC to CE differs across the three heterogeneous model pairs and varies during training.
Consequently, without ACR, the same fixed coefficient may produce different effective contributions of SCC, 
making it difficult to maintain a consistent balance with the task objective across model pairs and training stages.
Figure~\ref{fig:acr}(b) further shows that the SCC coefficient is regulated to different extents for the three model pairs 
according to their relative loss scales.
This regulation maintains adaptive control over the contribution of SCC despite variations in loss scale, 
thereby reducing sensitivity to coefficient selection.

\section{Conclusion}
In this work, redundancy suppression in heterogeneous knowledge distillation was revisited by considering the 
limitation of uniform decorrelation, which may weaken useful structural information encoded in feature correlations. 
CoCaRS addresses this limitation through semantic calibration of feature decorrelation while preserving cross-architecture 
invariance, together with adaptive regulation of the calibrated objective during optimization. 
Experimental results across heterogeneous model pairs consistently support the effectiveness of this formulation. 
Overall, these findings demonstrate the effectiveness of semantic correlation calibration in improving redundancy 
suppression for heterogeneous knowledge distillation.

\bibliography{aaai2027}

\end{document}